\documentclass[runningheads]{llncs}
\usepackage{graphicx}
\usepackage{subfig}
\usepackage{comment}
\usepackage{amsmath,amssymb} 
\usepackage{color}
\usepackage{hyperref}
\usepackage{breakurl}
\hypersetup{
    colorlinks=true,
    linkcolor=blue,
    filecolor=magenta,      
    urlcolor=cyan
}

\usepackage{multirow}
\usepackage{caption}

\begin{document}
\pagestyle{headings}
\mainmatter
\def\ECCVSubNumber{0000}  
\title{Ambient Sound Helps: Audiovisual Crowd Counting in Extreme Conditions} 

\titlerunning{Ambient Sound Helps: Audiovisual Crowd Counting in Extreme Conditions}
%

\author{Di Hu\inst{1}$^{,\star}$ \and
Lichao Mou\inst{2}$^{,\star}$ \and
Qingzhong Wang\inst{3}$^,$\thanks{The first three authors contribute equally to this work.} \and
Junyu Gao \inst{4} \and
Yuansheng Hua \inst{2,5} \and
Dejing Dou \inst{1} \and
Xiao Xiang Zhu \inst{2,5}
}

\authorrunning{D. Hu et al.}
%
\institute{Baidu Research
\email{\{hudi04,doudejing\}@baidu.com} \and
German Aerospace Center
\email{\{lichao.mou,yuansheng.hua,xiaoxiang.zhu\}@dlr.de} \and
City University of Hong Kong
\email{qingzwang2-c@my.cityu.edu.hk} \and
Northwestern Polytechnical University
\email{gjy3035@gmail.com} \and
Technical University of Munich
}

\maketitle

\begin{abstract}
Visual crowd counting has been recently studied as a way to enable people counting in crowd scenes from images. Albeit successful, vision-based crowd counting approaches could fail to capture informative features in extreme conditions, e.g., imaging at night and occlusion. In this work, we introduce a novel task of audiovisual crowd counting, in which visual and auditory information are integrated for counting purposes. We collect a large-scale benchmark, named auDiovISual Crowd cOunting (DISCO) dataset, consisting of 1,935 images and the corresponding audio clips, and 170,270 annotated instances. In order to fuse the two modalities, we make use of a linear feature-wise fusion module that carries out an affine transformation on visual and auditory features. Finally, we conduct extensive experiments using the proposed dataset and approach. Experimental results show that introducing auditory information can benefit crowd counting under different illumination, noise, and occlusion conditions. Code and data have been made available at \url{https://github.com/qingzwang/AudioVisualCrowdCounting}.
\end{abstract}

\section{Introduction}\label{sec1}

Crowd counting has recently been a hot research topic~\cite{babu2018divide,liu2018leveraging,shi2018crowd,shi2019revisiting}, as it can benefit a wide range of applications, to name a few, safety monitoring, public space design, and disaster management. Consequently, crowd counting techniques, particularly computer vision-based approaches, have received increased interest. The success of current state-of-the-art visual crowd counting models can be attributed to the development of convolutional neural network (CNN) architectures that aim at learning better visual representations from images for this task~\cite{li2018csrnet,liu2019context}. Albeit successful, vision-based crowd counting approaches could fail to capture informative features in extreme conditions\footnote{In this paper, the extreme condition refers to a) low resolution, b) noise, c) occlusion, and d) low illumination.} (c.f., Fig. 1).
\par
Investigations in the field of neurobiology show that human perception usually benefits from the integration of both visual and auditory information~\cite{stein1993merging}, e.g., lip reading, where correlations between lip movements and speech provide a strong cue for linguistic understanding~\cite{calvert1997activation}. This gives us an incentive that ambient sound could be an important cue for identifying the number of people in a scene. This hypothesis is in line with our daily experiences: \textit{the louder we perceive the ambient sound to be, the more people there are.} However, incorporating the ambient sound into a visual crowd counting model and its contributions to this task still remain underexplored in the community. On the other hand, with the now widespread availability of smartphones, digital cameras, and video surveillance equipments, audiovisual data have been accessible at a reasonable cost. This enables us to explore the topic in this paper.
\par
In this paper, we are interested in a novel task, audiovisual crowd counting. We pose and seek to answer the following questions:
\begin{itemize}
\item Is combining features coming from visual and auditory modalities better than only using visual features for crowd counting in extreme conditions?
\item How do audiovisual crowd counting results vary under different illumination, noise, and occlusion conditions?
\item How do we impose the audio information for effectively assisting the visual perception, i.e., how to fuse both modalities?
\begin{figure}[t!]
    \centering
    \includegraphics[width=1\textwidth]{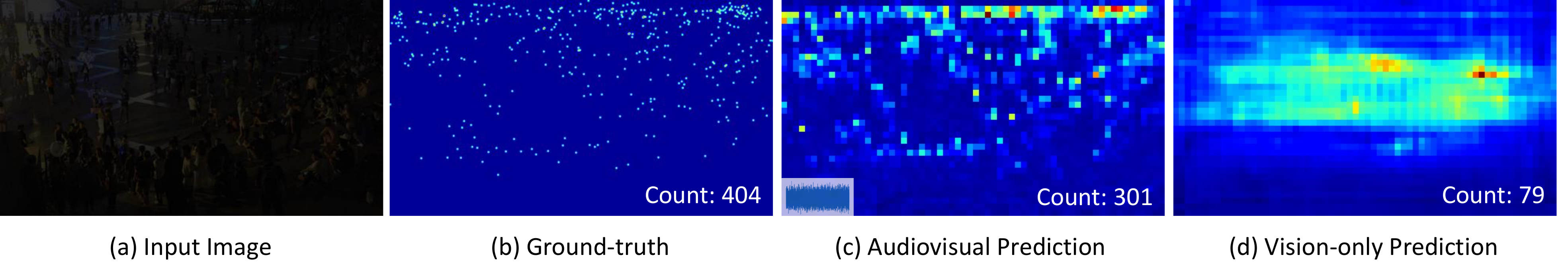}
    \caption{~Crowd counting on low-quality images. From left to right: input image with low illumination and strong noise, ground truth density map, predicted density map using both auditory and visual information and predicted density map only using visual information.\label{fig:eye-catch}}
\end{figure}%
\end{itemize}
\par
To this end, we establish an annotated benchmark, auDiovISual Crowd cOunting (DISCO) dataset. This allows for quantitative comparisons. As to the fusion of two modalities, inspired by FiLM \cite{perez2018film}, we also employ a feature fusion block where audio and visual features are fused in a linear manner. The advantages of linear feature fusion is its simplicity and less consumption of time and computational resources. To better fuse the two modalities, we stack multiple feature fusion blocks in our implementation (details are shown in Section~\ref{sec4}).
This work's contributions are threefold.
\begin{itemize}
\item Inspired by the audiovisual perception ability of human-beings, we investigate a novel audiovisual crowd counting task. We are not aware of any previous work exploring this topic. 
\item To facilitate progress in this field, we provide a large-scale benchmark, which consists of $1,935$ pairwise image-audio clips and $170,270$ annotated instances. This dataset covers a large variety of scenes in different illuminations.
\item We develop an audiovisual crowd counting model, based on which we carry out extensive experiments to explore in what situation and to what extent ambient sound benefits crowd counting, and achieve considerable audiovisual performance under extreme conditions.
\end{itemize}

\section{Related Work}
\noindent
\textbf{Crowd Counting.}
Early approaches \cite{chen2012feature,idrees2013multi,lempitsky2010learning,ryan2009crowd,sindagi2017generating,li2018csrnet,wang2019learning,liu2019context} mainly focus on estimating the number of people in crowd scenes with hand-crafted features, including Harr-like \cite{viola2004robust}, HOG \cite{dalal2005histograms} and so on. Recently, CNN-based methods \cite{cao2018scale,zhang2016single,sam2017switching,idrees2018composition,jiang2019crowd} have attained remarkable improvements by introducing architectures, such as multi-column, multi-stream, and trellis structure. Among these studies, \cite{li2018csrnet,liu2019context,sindagi2017generating,wang2019learning} aims at extracting global features or expanding the size of respective fields to encode the contextual information comprehensively. To fully mine the power of CNN, some methods propose novel feature fusion strategies (simple concatenation, element-wise sum operation \cite{liu2019crowd}, adversarial learning \cite{sindagi2017generating}, iterative CNN \cite{ranjan2018iterative}, bottom-top/top-bottom fusion \cite{sindagi2019multi}) to integrate multiple deep features from different layers or modules. 
\par
Although CNN-based approaches achieve a significant progress, there are two issues in the traditional single-image counting: 1) object occlusion results in missing estimation; 2) RGB sensors are susceptible to light intensity, object occlusion, visibility, etc. To address the first problem, some researchers \cite{zhang2019wide,maddalena2014people} propose multi-view methods to count the number of people. However, it is not easy to simultaneously carry multiple cameras with specific parameters and camera calibration in the real world. Regarding the second problem, although Lian \emph{et al.} \cite{lian2019density} attempt to solve it by introducing depth information, it is infeasible to tackle above issues at the same time.
\par
\noindent
\textbf{Joint Audiovisual Representation Learning.}
Joint audiovisual learning is expected to reward the learning model merits from both visual and auditory modalities. Early audiovisual researches mainly focus on speech recognition~\cite{dupont2000audio} as the visual message is considered to be free of audio noise and can provide complementary information in the noisy condition. A certain improvement usually can be achieved by jointly modeling the facial/mouth movements and corresponding audio signals~\cite{ngiam2011multimodal,Hu_2016_CVPR}. Similar phenomenons can also be found in other recognition tasks, such as affect recognition~\cite{zeng2007audio} and gesture recognition~\cite{kettebekov2005prosody}. Recently, audiovisual learning has been further employed to analyze more general scenarios. 
Owens \emph{et al.}~\cite{owens2016ambient} propose to transfer knowledge learned from audio modality to supervise the training of visual recognition models. Furthermore, Arandjelovic \emph{et al.}~\cite{arandjelovic2017look} propose to analyze video scenes with only audiovisual correlations, which could then be applied to sound localization~\cite{hu2019deep,hu2020curriculum} and separation~\cite{Gan_2019_ICCV}.
Inspired by these audiovisual works, we propose to transmit the merits of audio modality for crowd counting, which is considered to be free of visual noise and could provide considerable references.
\section{Dataset}
\begin{figure}[t]
    \centering
    \includegraphics[width=1\textwidth]{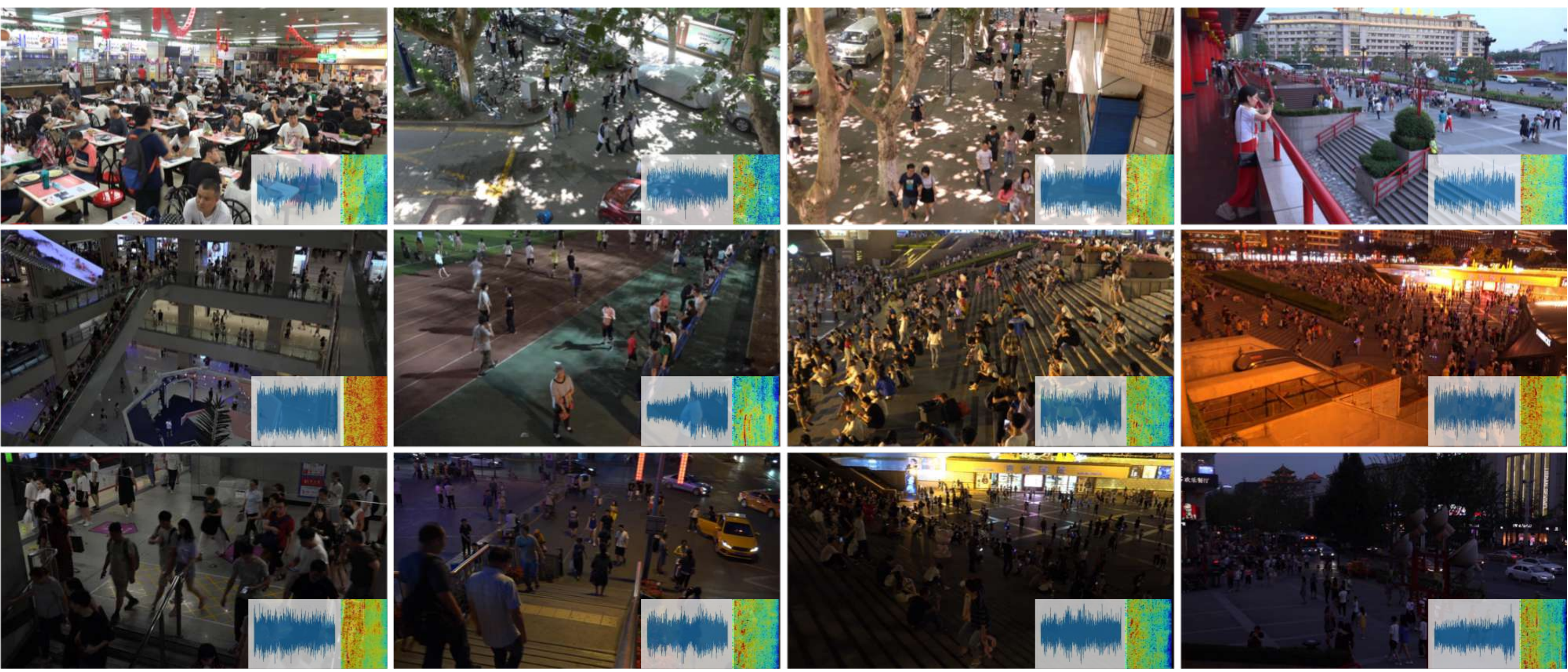}
    \caption{Examples of the DISCO dataset. We collect images and ambient sounds from a wide range of scenes, including indoor, outdoor, day- and night-time. For each image, an audio signal of one second is clipped from raw ambient sounds as auxiliary information. Raw ambient audio signals and their corresponding spectrograms are shown at the right bottom of each example.}
    \label{fig:data_examples}
\end{figure}



To jointly utilize ambient sounds and visual contexts for crowd counting, an auDIoviSual CrOwd dataset (``DISCO'' for short) is constructed. In this section, we will describe the proposed dataset from the following two perspectives: data collection and data characteristics. 

\textbf{Data Collection.} To simultaneously capture the visual image sequences and record the audio signals, we use four video cameras, HDR-CX900E produced by \emph{Sony Corporation}. In the collection process, we simulate the view of a surveillance camera and record crowds in some typical scenes at different time. As a result, we collect $248$ video clips, around $20$ hours and $385$ GB data in total. Specifically, the resolution of each video is  $1,920\times1,080$, and the frame rate is $25$. For the audio information, the DV record 2-channel stereo with the sample rate of $48,000$. 

From these raw data, 1,935 images and audios from various typical scenes are selected to construct our proposed dataset. For an image at $t$ in a video, we extract its corresponding audio signals from $t-0.5s$ to $t+0.5s$. Some visual examples and their corresponding audio waveforms are shown in Fig. \ref{fig:data_examples}. 

\begin{figure}[t]
\centering
\begin{minipage}[b]{.4\linewidth}
    \centering
    \includegraphics[width=\textwidth]{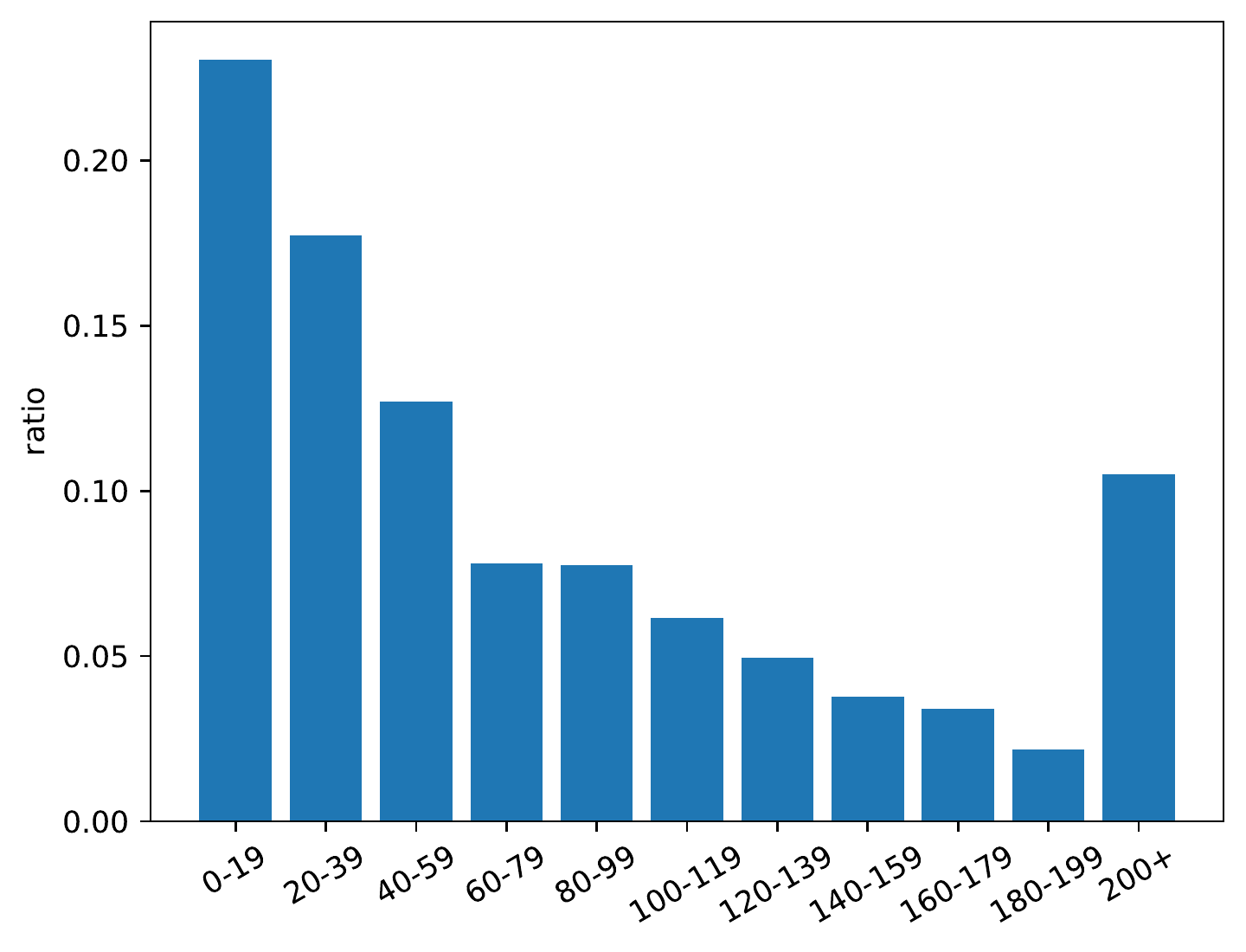}
\end{minipage}
\begin{minipage}[b]{.58\linewidth}
    \centering
    \includegraphics[width=\textwidth]{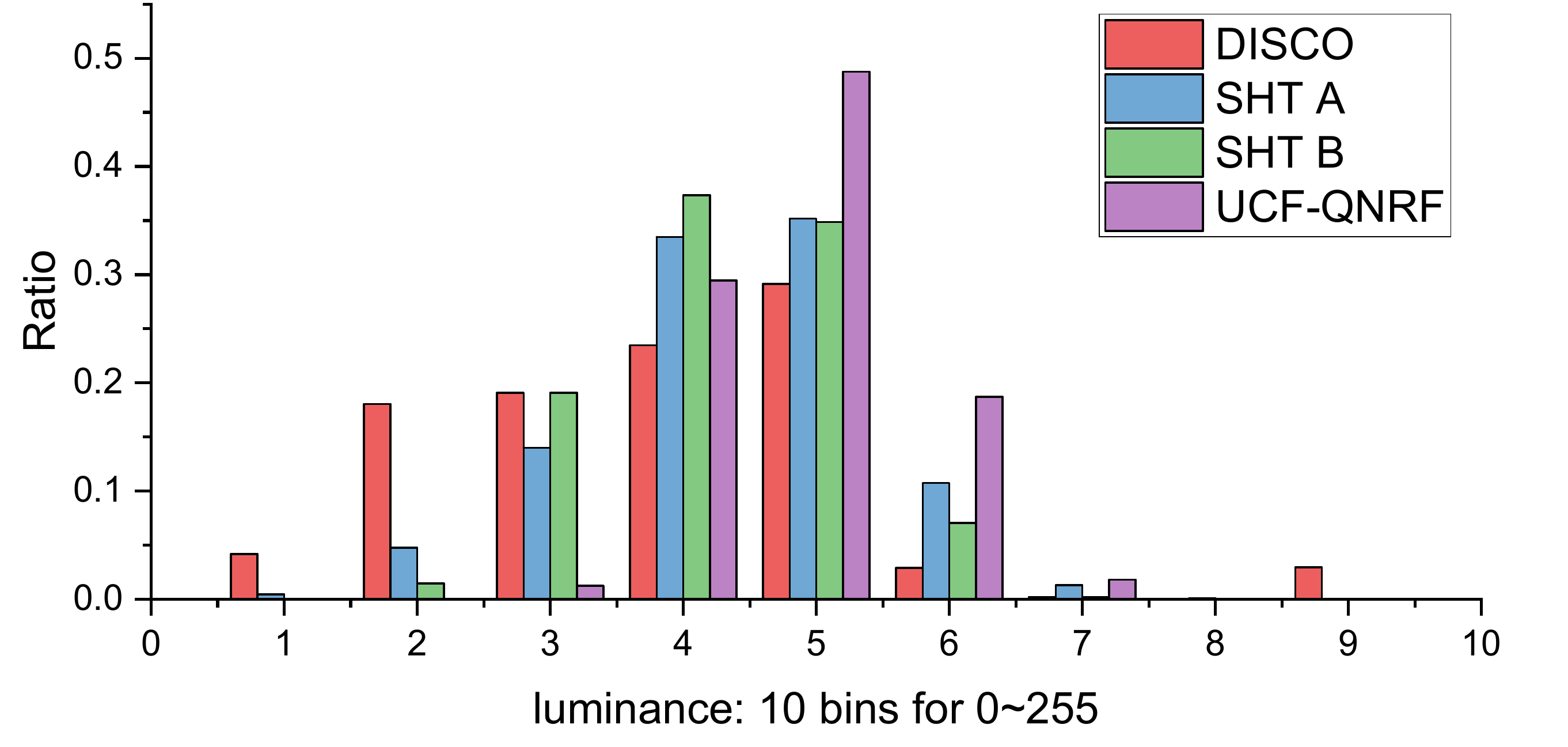}
\end{minipage}
\caption{Left: the statistical histogram of counts on the DISCO dataset. Right: comparisons of illumination distribution among different datasets.}\label{Fig-hist}
\end{figure}

\textbf{Data Characteristics.} DISCO consists of $1,935$ crowd images, a total of $170,270$ instances annotated with the head locations. The average, minimum and maximum number of people for each image are $87.99$, $1$ and $709$, respectively. Fig. \ref{Fig-hist} reports the histogram of the population distribution. From this, we find that the number of people in most scenes is between $0$ and $100$. 
 
Compared with some traditional crowd counting datasets \cite{zhang2016single,idrees2018composition}, the proposed DISCO dataset are the first to record ambient sounds as auxiliary information of crowd scenes to reduce defects of single-vision sensors. In addition, we capture images at different times in one day to ensure their various illuminations (see Fig. \ref{fig:data_examples}). As illustrated in Fig. \ref{Fig-hist}, We also analyze the illumination distribution in YUV space of DISCO and compare it with Shanghai Tech Part A/B (SHT A, SHT B for short) \cite{zhang2016single} and UCF-QNRF \cite{idrees2018composition}. It can be observed that the illumination of ~25\%/3\% of images in DISCO dataset is extremely low/high (see the first two and 9th bins), while other datasets rarely involve very few images of such poor quality. What's more, different from traditional video-surveillance-style datasets \cite{chan2008privacy,chen2012feature,zhang2016data}, the proposed DISCO dataset covers various scenes from different cities, such as subway station, mall, restaurant, campus, plaza, stadium, sidewalk, etc.

In a summary, DISCO dataset has three advantages comparing with others: 1) both audio and visual signals are provided; 2) cover different illuminations; and 3) a large variety of scenes are considered. 

\section{Our Approach}\label{sec4}

\subsection{Overview}
\begin{figure*}[t]
    \centering
    \includegraphics[width=1\textwidth]{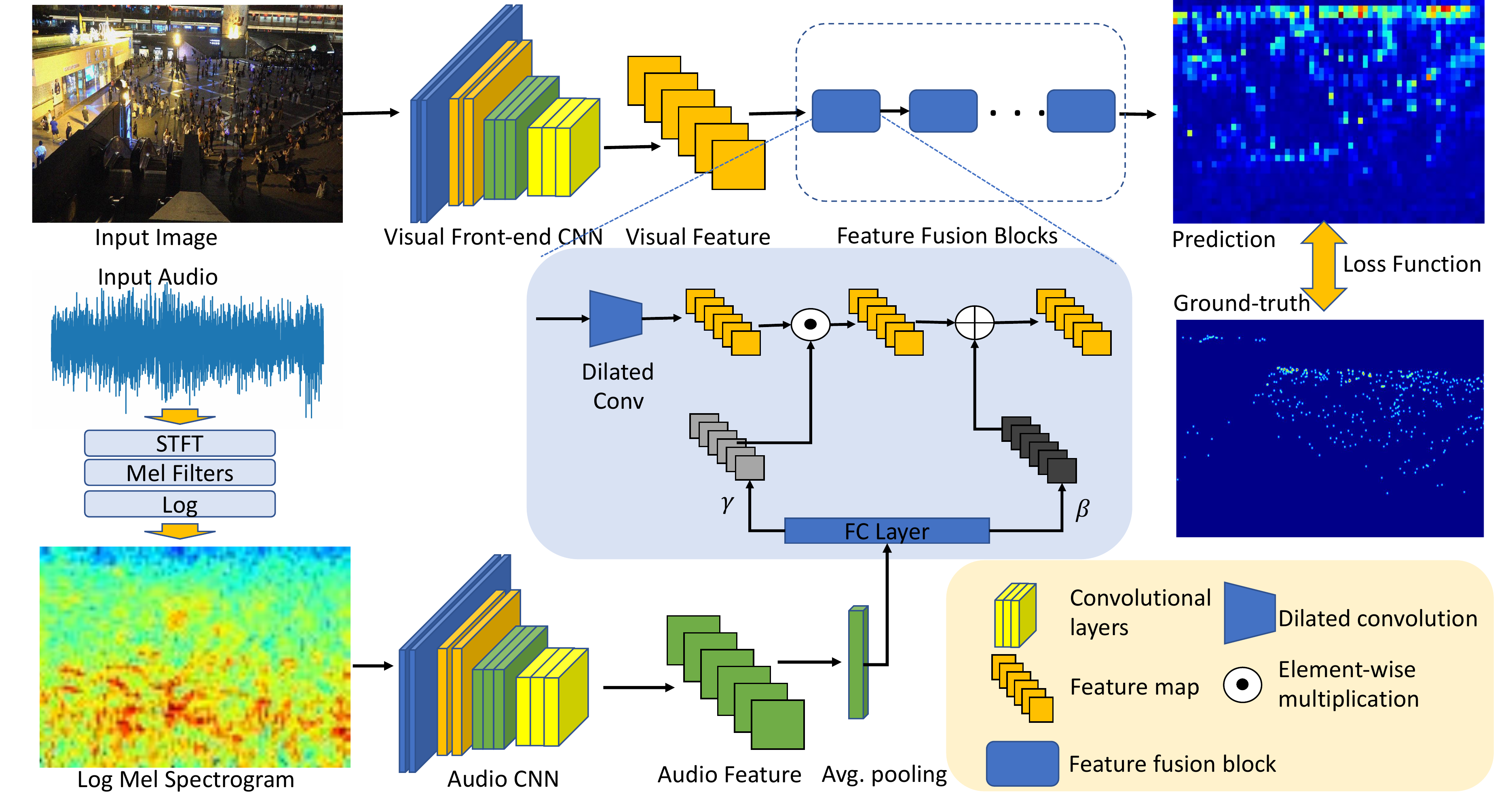}
    \caption{Overview of the proposed AudioVisual Counting model (AVC). The proposed AVC model is composed of three modules (1) visual feature extraction, (2) audio feature extraction,  and (3) feature fusion. Note that our AudioVisual Counting framework can be applied to any vision-based counting model.}
    \label{fig:model_overview}
\end{figure*}
In order to benefit crowd counting with ambient sounds, a novel AudioVisual Counting (AVC) network is designed and consists of three modules (see Fig. \ref{fig:model_overview}): (1) visual feature extraction, (2) audio feature extraction, (3) audiovisual feature fusion. Notably, comparing with traditional methods \cite{cao2018scale,idrees2018composition,jiang2019crowd,zhang2016single,sam2017switching}, where only visual information is employed, our network is characterized by the second and third module. As aforementioned, one of our motivations is that in the scenario of low-quality images, such as low resolution, low illumination and strong noise, it could be difficult to precisely count people with only visual information. In contrast, humans is able to estimate the number of people, even though we cannot see them, the reason is that we are able to perceive the scene by hearing. To imitate such human capacity, we introduce an audio module into the traditional counting framework, resulting in {\it AudioVisual Counting (AVC)} model.

\subsection{Visual Feature Extraction}
Similar to CSRNet \cite{li2018csrnet}, we employ the first ten layers of VGG16 \cite{simonyan2014very} as the front-end CNN $\mathcal{V}_{CNN}$ to extract visual features. Given an RBG image $I$ with a spatial size of $W \times H$, visual features $v_{feat}$ can be extracted with the following equation:
\begin{equation}
    v_{feat} = \mathcal{V}_{CNN}(I),
\end{equation}
where $v_{feat}\in \mathbb{R}^{C\times\frac{W}{8}\times\frac{H}{8}}$, and $C$ denotes the number of channels, i.e., 512.

\subsection{Audio Feature Extraction}
In this work, we use Log Mel-Spectrogram (LMS) for representing audio and CNN arch for modeling due to following considerations: 1) The audio feature of LMS has been widely used in CNN-like neural model for sound event detection and shown noticeable performance\cite{hershey2017cnn}, and  2)  Stoter et al.\cite{stoter2018classification} demonstrates that using spectrogram-like feature can achieve comparable performance to the conventional MFCC in the counting task and much simpler. Even so, we still provide some discussions about different audio features and modeling settings in the experiments.

Given a raw audio signal $A_{raw}=\{a_1, a_2, \cdots, a_T\}$, we first sub-sample $A_{raw}$ at 16kHz, and then employ short-time Fourier transform (STFT) using Hann window with the window size of 400 and a hop length of 160, to generate a $98\times 257$ time-frequency map. Afterwards, Mel filter bank is applied, and a $96\times 64$ representation $A_{spec}$ can be then obtained for each raw audio signal. Finally, we utilize a VGG-like deep convolutional neural network \cite{hershey2017cnn} to extract audio features $a_{feat}$ as follows:
\begin{equation}
    a_{feat}=\mathcal{A}_{CNN}(A_{spec}),
\end{equation}
where $a_{feat}\in \mathbb{R}^{C\times W_a\times H_a}$, and $C=512$.

\subsection{Feature-wise Audiovisual Fusion}
To effectively fuse both audio and visual information in crowd counting, we introduce a feature-wise fusion module which aims at adaptively adjusting visual feature responses with transformed audio embeddings.
Concretely, based on the extracted audio features, two feature-wise parameters $\gamma$ and $\beta$  are learned to model such cross-modal influence in terms of multiplicative and additive aspects, respectively. The formula is shown here:
\begin{equation}\label{fusion}
v_{l+1} = \mathcal{F}_l\left(\gamma_{l}\odot \mathcal{D}_{CNN}^l(v_l) + \beta_l\right),
\end{equation}
where $v_{l}\in\mathbb{R}^{C_l\times W_l\times H_l}$ indicates outputs of the $l$th feature fusion block, $\mathcal{D}_{CNN}^l$ denotes the $l$th dilated convolution layer, $\mathcal{F}_l$ and $\odot$ suggest the activation function and element-wise multiplication, respectively. Notably, $l$ ranges from 0 to 6, and $v_0 = v_{feat}$. Normally, $\gamma$ and $\beta$ can be learned via different affine transformations, such as single or multiple neural networks. In this work, we simply use 
fully-connected layers to learn $\gamma_l$ and $\beta_l$ with the following two equations:
\begin{align}
    \gamma_l &= FC_l^{\gamma}(AvgP(a_{feat})),\\
    \beta_l &= FC_l^{\beta}(AvgP(a_{feat})).
\end{align}
In these two equations, $AvgP$ represents average pooling, and $\gamma, \beta \in \mathbb{R}^{C_{l+1}}$. To implement Eq.\ref{fusion}, $\gamma_l$ and $\beta_l$ are tiled to match the size of visual features before fusion, see Fig.\ref{fig:model_overview}.  

Intuitively, the feature-wise fusion module manipulates visual feature maps by referring to their corresponding ambient sound information independently, especially when faced with low-quality images. Note that, as $\gamma$ and $\beta$ are irrelevant to the spatial dimension of visual features, the fusion module can be applied after arbitrary visual feature maps, which makes it possible to further improve the fusion performance by inserting the fusion module at different levels simultaneously. In this condition, we can also share the same affine transformation in different fusion modules for an efficiency purpose, as shown in Fig.\ref{fig:model_overview}.

\subsection{Loss Function}
Given a ground-truth density map $Y\in \mathbb{R}^{W\times H}$ and a predicted density map $\hat{Y}\in \mathbb{R}^{W\times H}$, we select $L_2$ norm as the loss function, and the loss can be calculated with the following equation:
\begin{equation}
\mathcal{L}=\sum_{i=1}^W\sum_{j=1}^H \left(Y_{ij}-\hat{Y}_{ij}\right)^2.
\end{equation}


\section{Experiments}
In this section we present the settings, results, and analysis of the experiments. More results and discussions are in the Appendix.

\subsection{Experimental Settings}
First, we split our DISCO dataset into three sets: 200 images for validation, 300 images for testing, and the remaining 1,435 images for training. To obtain the ground-truth density maps, we convolve each binary annotations (centers of human heads are one, and the others are zero) with a $15\times 15$ Gaussian kernel $\mathcal{K}\sim \mathcal{N}(0, 4.0)$.

In the training phase, we select Adam \cite{kingma2014adam} as the optimizer and set its parameters as recommended. The learning rate is initialized as $1e-5$ and decays by 0.99 every epoch. To alleviate overfitting, weight decay is employed with a $\lambda$ of $\text{1e-4}$. It is noteworthy that except for those with a low resolution of $128 \times 72$, we resize images into $1024\times 576$ to reduce computational resources and time. In our experiments, the batch size is set as 4, and the maximum training epoch is 500. At the end of each epoch, models are evaluated on the validation split, and only those of the best performances are remained after training. To fairly compare all models, we report their performances on the test split.

Regarding the audio CNN, we use VGGish \cite{hershey2017cnn} pre-trained on audioSet \cite{audioset} and discard its last three  fully-connected layers, resulting in a 6-layer CNN. For the visual front-end CNN, we use the first ten layers of VGG16 \cite{simonyan2014very} pre-trained on ImageNet \cite{ILSVRC15}. 

In our feature fusion blocks, we employ dilated convolutions \cite{yu2015multi} with the kernel size of 3 and the dilation rate of 2 to enlarge the receptive field. Similar to CSRNet \cite{li2018csrnet}, we stack 6 fusion blocks and up-sample outputs with a factor of 8 to yield density maps of full resolution.

\subsection{Baselines and Evaluation Metrics}
To investigate the task of audiovisual crowd counting, we compare our audiovisual counting model with several vision-based models, such as MCNN \cite{zhang2016single}, CSRNet \cite{li2018csrnet}, SANet \cite{cao2018scale}, and CANNet \cite{liu2019context}. Notably, we use one of the  state-of-the-art models, CSRNet \cite{li2018csrnet}, as the backbone of our audiovisual counting model, leading to the proposed AudioCSRNet. The architecture of our proposed network is shown in Fig. \ref{fig:model_overview}.

To assess the performance of each model, we employ \textbf{Mean Absolute Error (MAE)} and \textbf{Mean Square Error (MSE)} scores, which are computed as follows:
\begin{equation}
\text{MAE}=\frac{1}{N}\sum_{n=1}^N|c_n-\hat{c}_n|,~~~\text{MSE} = \sqrt{\frac{1}{N}\sum_{n=1}^{N}|c_n-\hat{c}_n|^2},
\end{equation}
where $N$ denotes the number of images, $c$ and $\hat{c}$ denote the ground-truth and predicted numbers of people.

\subsection{Audiovisual Crowd Counting on Low-quality Images}
\newcommand{\tabincell}[2]{\begin{tabular}{@{}#1@{}}#2\end{tabular}}
\begin{table}[t]
    \centering
    \caption{Performance on low-quality images. For Gaussian noise, the standard deviation denoted by $\sigma$ is a fixed value. While in low illumination\&Gaussian noise, the illumination decay rate $r$ and standard deviation $\sigma$ of Gaussian noise are random values and $R$ and $B$ represent the hyper-parameters to compute $r$ and $\sigma$, respectively. The bold numbers denote the best performance and the blue numbers represent the second best performance.}
    \label{tab:lowq_img}
    \scalebox{0.75}{
    \begin{tabular}{c|c|c|c|c|c|c|c|c|c|c|c|c}
        \hline
         \multirow{3}{*}{\tabincell{c}{Model \& \\ Image Quality}} &\multicolumn{2}{c|}{Low resolution} &\multicolumn{4}{c|}{Gaussian noise} &\multicolumn{4}{c|}{Low illumination\&Gaussian noise}  &\multicolumn{2}{c}{\multirow{2}{*}{\textbf{Avg. Score}}}\\
         \cline{2-11}
         &\multicolumn{2}{c|}{$128\times 72$} &\multicolumn{2}{c|}{$\sigma=25/255$} &\multicolumn{2}{c|}{$\sigma=50/255$} &\multicolumn{2}{c|}{$R=0.2, B=25$} &\multicolumn{2}{c|}{$R=0.2, B=50$} \\
         \cline{2-13}
         &MAE $\downarrow$ &MSE $\downarrow$ &MAE $\downarrow$ &MSE $\downarrow$ &MAE $\downarrow$ &MSE $\downarrow$ &MAE $\downarrow$ &MSE $\downarrow$ &MAE $\downarrow$ &MSE $\downarrow$ &MAE $\downarrow$ &MSE $\downarrow$ \\
         \hline
         MCNN \cite{zhang2016single} &60.17 &89.35 &53.47 &84.04 &53.92 &84.04 &70.72 &96.11 &70.58 & 96.11 &61.77 &89.93 \\
         \hline
         CANNet \cite{liu2019context} &22.16 &39.60 &\textcolor{blue}{13.31} &\textbf{27.23} &\textcolor{blue}{14.20} &\textbf{28.04} & \textcolor{blue}{26.03} &\textbf{49.11} &\textcolor{blue}{33.14} &\textcolor{blue}{58.27} &\textcolor{blue}{21.77} &\textcolor{blue}{40.45} \\
         \hline
         CSRNet \cite{li2018csrnet} &\textcolor{blue}{17.14} &\textbf{30.64} &13.79 &28.01 &14.55 &29.15 &35.78 &62.76 &45.88 &75.40 &25.43 &45.19 \\
         \hline
         AudioCSRNet &\textbf{16.88} &\textcolor{blue}{31.46} &\textbf{13.07} &\textcolor{blue}{27.45} &\textbf{13.70} &\textcolor{blue}{28.67} &\textbf{25.06} &\textcolor{blue}{51.58} &\textbf{27.33} &\textbf{45.16} &\textbf{19.21} &\textbf{36.86}\\
         \hline
         \hline
         PSNR \cite{hore2010image} $\uparrow$ &\multicolumn{2}{c|}{22.27} &\multicolumn{2}{c|}{30.05} &\multicolumn{2}{c|}{24.13} &\multicolumn{2}{c|}{9.94} &\multicolumn{2}{c|}{10.43} & \multicolumn{2}{c}{---}\\
         \hline
         BRISQUE \cite{mittal2012no} $\uparrow$ &\multicolumn{2}{c|}{29.75} &\multicolumn{2}{c|}{82.19} &\multicolumn{2}{c|}{69.06} &\multicolumn{2}{c|}{56.08} &\multicolumn{2}{c|}{66.39} & \multicolumn{2}{c}{---}\\
         \hline
    \end{tabular}
    }
\end{table}

\begin{figure}[t]
\centering
\begin{tabular}{cc}
$B=25$ &$B=50$ \\
\includegraphics[width=0.45\textwidth]{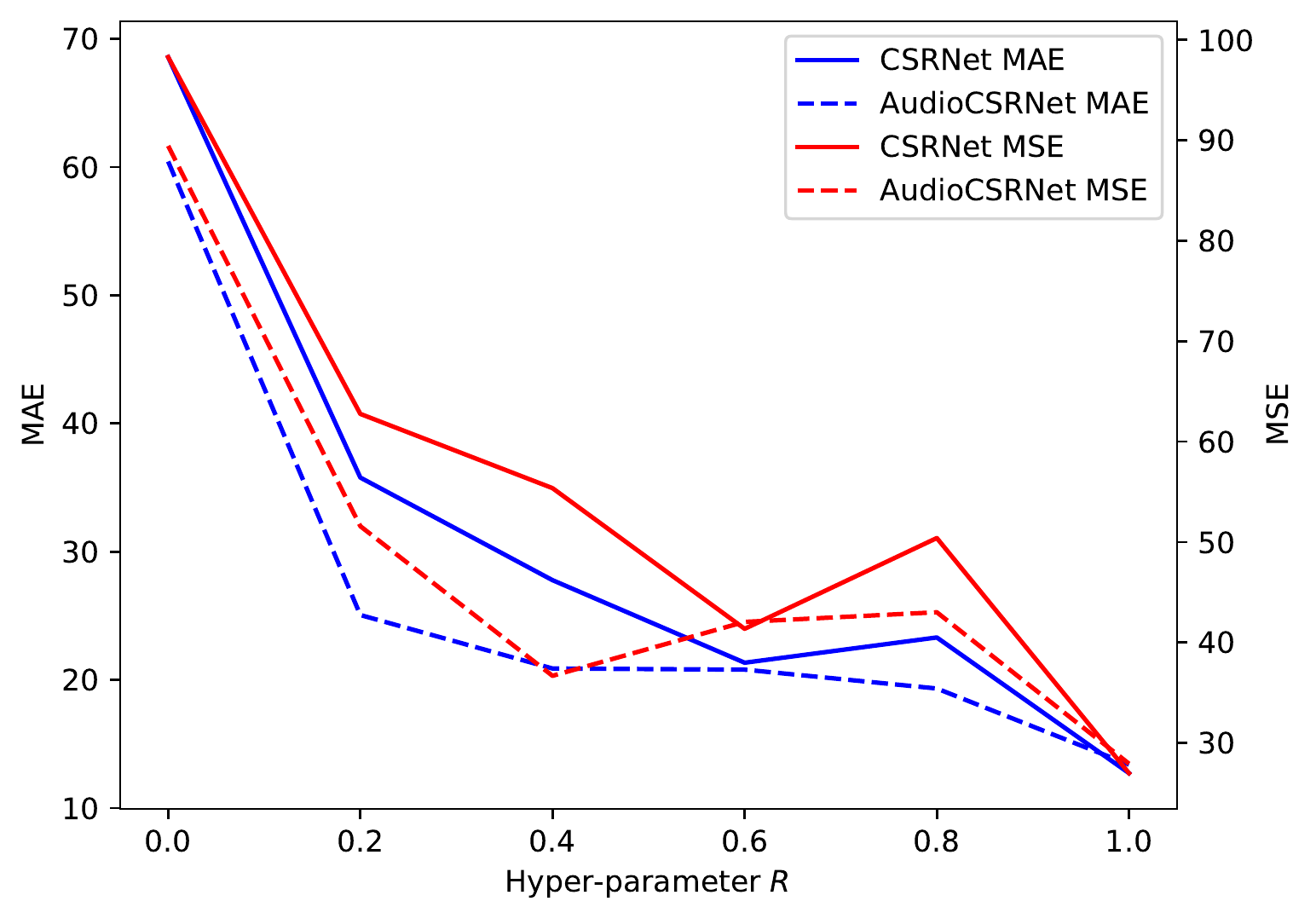} &\includegraphics[width=0.45\textwidth]{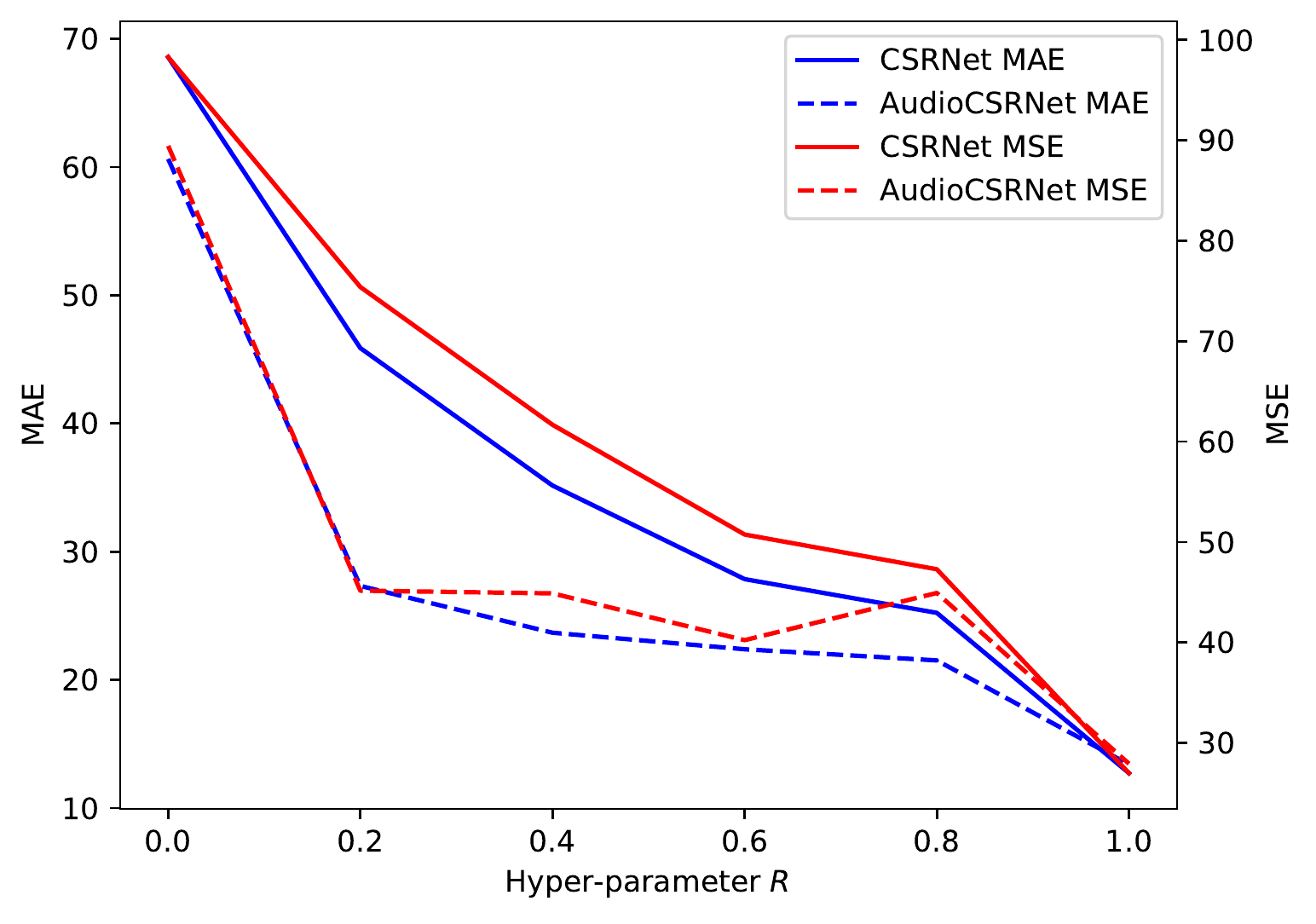}
\end{tabular}
\caption{The effect of different illuminations. $R=0$ means that there is no visual information and $R=1.0$ means to use the original image without lowering the brightness and adding Gaussian noise.}\label{diff-illumination}
\end{figure}

\begin{figure*}[t]
    \centering
    \includegraphics[width=.9\textwidth]{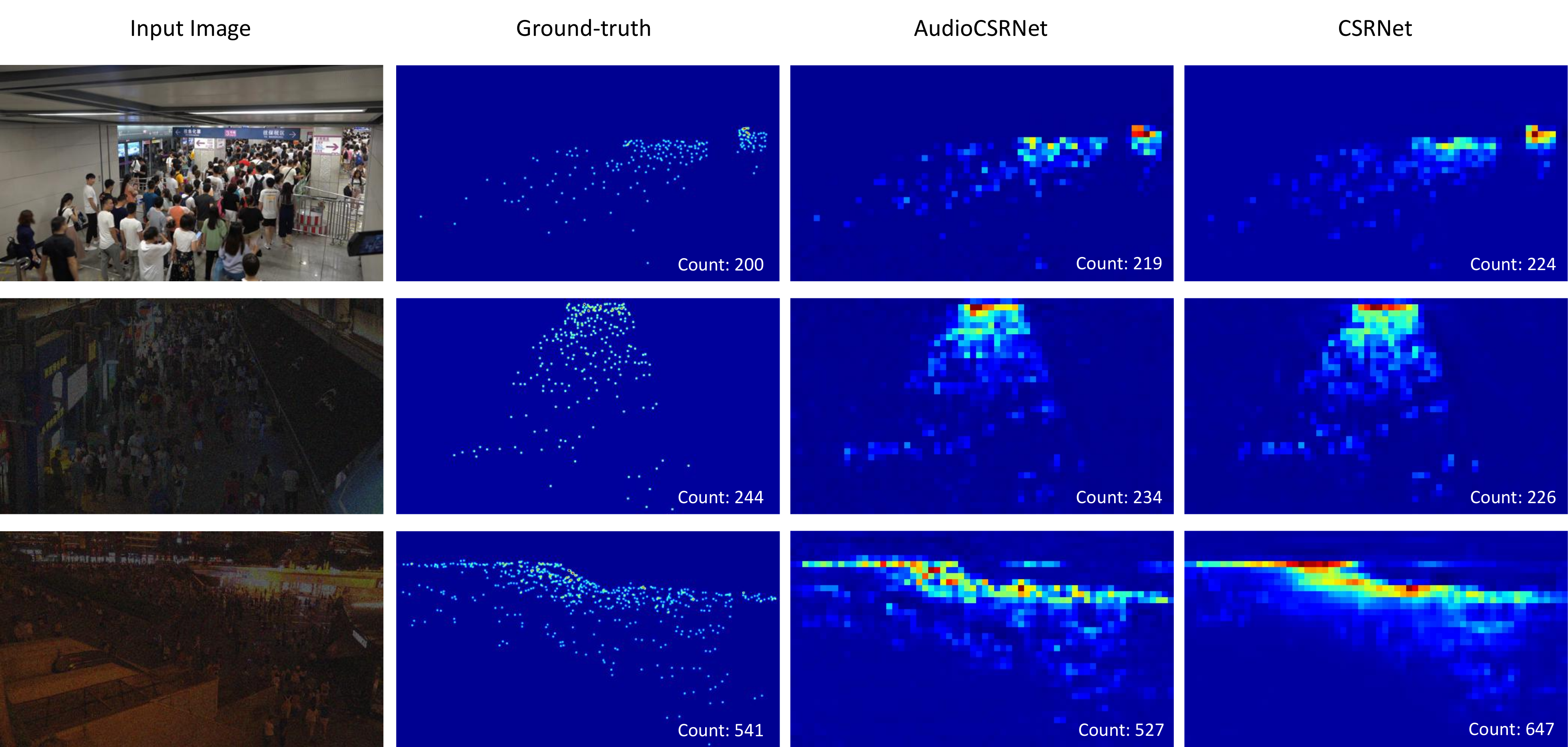}
    \caption{Density maps generated by CSRNet and AudioCSRNet on low-quality images.}
    \label{density_map_example}
\end{figure*}

To evaluate the performance of our AudioCSRNet in crowd counting under extreme conditions, we conduct experiments on two extreme scenarios: 1) the quality of images is very low, and 2) occlusion exists in images. In this subsection, we focus on the first scenario and test our model on three low-quality conditions: low illumination, low resolution, and strong noise. Specifically, we mimic images taken in the dark environment\footnote{The problem of directly using natural extreme low-illumination images is that the annotation of human heads in these images is difficult owing to insufficient viual information. To solve this, we first label images with low illumination and then produce pseudo extreme low-illumination by imitating the extreme scenario.} with the method proposed by \cite{lore2017llnet}. First, we randomly reduce the brightness of images with a rate of $r$, calculated by $R*\textbf{Uniform}(0, 1)$. Then we add Gaussian noise to the low illumination image and the standard deviation $\sigma$ of the Gaussian noise is a random variable, i.e., $\sigma=\sqrt{\textbf{Uniform}(0, 1)*\left(\frac{B}{255}\right)^2}$, where $R$ and $B$ are two hyper-parameters to control brightness decay rate and the variance of Gaussian noise. To quantitatively measure the quality of input images, here we calculate PSNR \cite{hore2010image} and BRISQUE \cite{mittal2012no}. Notably, images with high PSNR and BRISQUE scores are regarded as high-quality ones.\footnote{This is slightly different with the original BRISQUE score. Normally, lower BRISQUE scores indicate higher-quality images, but here we use 100-BRISQUE, thus, higher scores indicate higher-quality images.}.

Table \ref{tab:lowq_img} shows comparisons among different models. 
Specifically, comparisons between AudioCSRNet and its counterpart, CSRNet, directly demonstrate that introducing audio information can benefit crowd counting, in particular on lower-quality images. For example, on images with low illumination and Gaussian noise (PSNR is 9.94 and 10.43), CSRNet obtains the MAEs of 35.78 and 45.88, whereas the MAEs of AudioCSRNet drop to 25.06 and 27.33, respectively. Besides, on images with low resolution (PSNR is 22.27), AudioCSRNet surpasses all competitors and achieves decrements of up to 43.29 in terms of MAE. As to images corrupted with Gaussian noise (PSNR is 30.05 and 24.13), AudioCSRNet obtains the lowest MSE in comparison with visual models as well.


Another advantage of introducing audio into crowd counting is that audiovisual models show strong robustness on variant scenarios, e.g., AudioCSRNet obtains the lowest average MAE and MSE score (19.21 and 36.86). \textcolor{black}{According to the no free lunch theorem \cite{wolpert1997no}---no model is able to outperform the others in all scenarios, while introducing audio to crowd counting balances the performance and in most cases audiovisual models performs better than models using only visual information.
}

To further evaluate the effectiveness of AudioCSRNet, we conduct ablation studies with different illumination reduction strategies, as shown in Fig. \ref{diff-illumination}. Generally, AudioCSRNet is able to obtain lower MAE and MSE score when the illumination is reduced gradually. An interesting observation is that when $R$ is $0$, which indicates that nothing can be observed, AudioCSRNet still works and can achieve an MAE of 60.43 and MSE of 89.45. This means that even with only ambient sounds, AudioCSRNet can count people. This is in line with the reality that human can roughly estimate the number of people with only auditory information. Although the number estimated by hearing is not precise, low-quality images can be used to calibrate it. For instance, when $R=0.2, B=25$, MAE and MSE obtained by AudioCSRNet decrease to 25.06 and 51.58, respectively, which is much lower than those obtained by CSRNet (35.78 and 62.76). By increasing the intensity of Gaussian noise, such difference enlarges as well, i.e., the MAE of AudioCSRNet is decreased by 18.55 compared to that of CSRNet, when $R=0.2, B=50$.

Also, we show the density map predicted by CSRNet and AudioCSRNet in Fig.~\ref{density_map_example}. Jointly applying auditory and visual information leads to more accurate density maps, especially for images that suffers from stronger noise and lower illumination. In Fig.~\ref{density_map_example}, the first row shows the comparison on a high quality image, where density maps predicted by AudioCSRNet and CSRNet are quite similar, while in the last row, the prediction of AudioCSRNet is superior to that of CSRNet when image quality decreases. To summarize, introducing audio information can effectively improve the performance of crowd counting with low-quality images

\subsection{Audiovisual Crowd Counting under Occlusion}

\begin{figure}[t]
\centering
\includegraphics[width=0.48\textwidth]{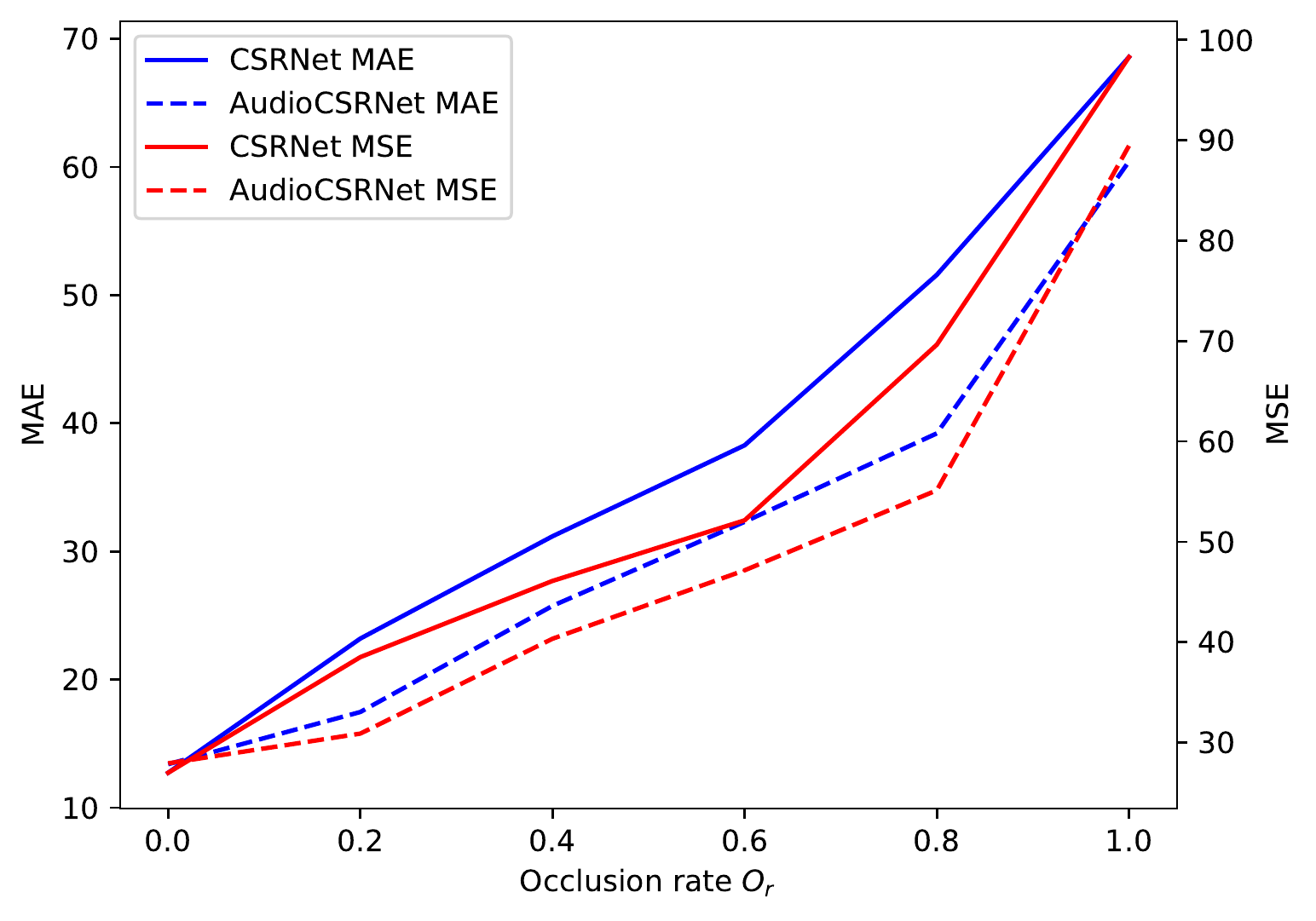}
\caption{Performance of CSRNet and AudioCSRNet on occluded images. Occlusion rate $O_r=0$ represent the original images and $O_r=1.0$ means that there is no visual information.}\label{occ-csrnet}

\end{figure}

Another scenario that we study is occlusion, where an input image is randomly occluded by a black rectangle. Specifically, given an image $I$ with the size of $W\times H$ and an occlusion rate $O_r$, we first generate a black rectangle with the size of $\textbf{int}(W\sqrt{O_r})\times \textbf{int}(H\sqrt{O_r})$, and then randomly mask out certain image contents with this black rectangle.

Results are shown in Fig. \ref{occ-csrnet}, and we can see that performances of both CSRNet and AudioCSRNet on occluded images dramatically decrease, e.g., CSRNet and AudioCSRNet achieve MAE scores of 23.20 and 17.47, respectively, when 20\% of input image area is occluded, which is 1.8 and 1.3 larger than the MAEs obtained on fully visible images. Moreover, trends of curves in Fig. \ref{occ-csrnet} demonstrate that AudioCSRNet can often achieve lower MAE and MSE scores, and with an increasing $O_r$, the gap between MAE and MSE scores achieved by CSRNet and AudioCSRNet grows as well, meaning that the changes of images can more easily affect CSRNet that only depends on visual information, while AudioCSRNet is more stable when the quality of images varies. 

\subsection{Vision vs. Audiovision on High-quality Images}

\begin{table}[t]
    \centering
    \caption{Performance on high-quality images. The bold numbers denote the best performance and the blue numbers represent the second best performance.}
    \label{tab:clean_img}
    \begin{tabular}{c|c|c|c}
        \hline
        Model & MAE $\downarrow$ &MSE $\downarrow$ &BRISQUE \cite{mittal2012no} $\uparrow$ \\
        \hline
        MCNN \cite{zhang2016single} &53.40 &84.10 &\multirow{5}{*}{78.03} \\
        \cline{1-3}
        SANet \cite{cao2018scale} &17.22 &32.00 \\
        \cline{1-3}
        CANNet \cite{liu2019context} &15.41 &28.96 \\
        \cline{1-3}
        \cline{1-3}
        CSRNet \cite{li2018csrnet} &\textbf{13.88} &\textcolor{blue}{28.79} \\
        \cline{1-3}
        AudioCSRNet &\textcolor{blue}{14.24} &\textbf{28.07} \\
        \hline
    \end{tabular}
\end{table}

Despite the success of AudioCSRNet in crowd counting under extreme conditions, here we are interested in whether ambient sounds help when high-quality images are available as well. Table \ref{tab:clean_img} shows comparisons between different models on high-quality images. Obviously, performances of all models on high-quality images increases compared to those on low-quality images (see table \ref{tab:lowq_img}). As shown in Table \ref{tab:clean_img}, CSRNet outperforms all other models and achieves 13.88 in terms of MAE. Besides, CANNet performs worse than CSRNet, and gains increments of 1.53 and 0.17 in MAE and MSE, respectively. Moreover, we find that the performance of AudioCSRNet is comparable to that of CSRNet, which achieves slightly worse MAE score (14.24), but better MSE score (28.07), and this interesting observation demonstrates that the introduction of audio information might not hurt the performance on high-quality images. One possible explanation could be that high-quality images (BRISQUE score reaches 78.03) can provide sufficient visual cues for the precise estimation of crowd counts. Notably, although images in our dataset cover a wide range of the illumination (see Fig. \ref{Fig-hist} right), those with extremely low illumination are not used in the annotation procedure. As a consequence, people are recognizable in all collected images. In this case, the application of audio information might bring noise and slightly reduce the network performance. 

By comparing experimental results on high-quality image with low resolution images (see table~\ref{tab:lowq_img}), it can be observed that the MAEs obtained by CSRNet and CANNet decrease by $19\%$ and $30.5\%$, respectively. This is because CANNet learns and aggregates features of different scales with multi-scale structures, and those extracted from low resolution images might introduce noise. An intuitive explanation could be that a person in a low resolution image consists of only a few pixels, and a kernel with the large receptive field might take irrelevant visual information, e.g., noise or background, into consideration when counting people.


\subsection{Audio Net as a General Module}

\begin{table}[t]
    \centering
    \caption{Performance on low-quality images w.r.t. CANNet~\cite{liu2019context} and corresponding audiovisual one AudioCANNet.}
    \label{tab:can_img}
    \scalebox{0.75}{
    \begin{tabular}{c|c|c|c|c|c|c|c|c|c|c|c|c}
        \hline
         \multirow{3}{*}{\tabincell{c}{Model}} &\multicolumn{2}{c|}{Low resolution} &\multicolumn{4}{c|}{Gaussian noise} &\multicolumn{4}{c|}{Low illumination\&Gaussian noise}  &\multicolumn{2}{c}{\multirow{2}{*}{\textbf{Avg. Score}}}\\
         \cline{2-11}
         &\multicolumn{2}{c|}{$128\times 72$} &\multicolumn{2}{c|}{$\sigma=25/255$} &\multicolumn{2}{c|}{$\sigma=50/255$} &\multicolumn{2}{c|}{$R=0.2, B=25$} &\multicolumn{2}{c|}{$R=0.2, B=50$} \\
         \cline{2-13}
         &MAE $\downarrow$ &MSE $\downarrow$ &MAE $\downarrow$ &MSE $\downarrow$ &MAE $\downarrow$ &MSE $\downarrow$ &MAE $\downarrow$ &MSE $\downarrow$ &MAE $\downarrow$ &MSE $\downarrow$ &MAE $\downarrow$ &MSE $\downarrow$ \\
         \hline
         CANNet \cite{liu2019context} &22.16 &39.60 &\textbf{13.31} &\textbf{27.23} & 14.20 &\textbf{28.04} & 26.03 &49.11 & 33.14 &58.27 & 21.77 & 40.45 \\
         \hline
         AudioCANNet &\textbf{16.88} &\textbf{32.24} &13.53 &27.88 &\textbf{13.69} &28.10 &\textbf{25.93} &\textbf{48.65} &\textbf{30.83} &\textbf{54.54} & \textbf{20.17} & \textbf{38.28} \\
         \hline
    \end{tabular}
    }
\end{table}


Note that our audio net shown in Fig. \ref{fig:model_overview} can also be applied to any other vision-based models. In this subsection we directly applied the audio net to CANNet, resulting in AudioCANNet, where the backbone is the first 10 layers of VGG16 and the scale-aware layer.

As shown in Table~\ref{tab:can_img}, compared with CANNet, AudioCANNet obtains lower MAE and MSE on low resolution images, i.e., AudioCANNet obtains 16.88 for MAE and 32.24 for MSE, whereas CANNet obtains 22.16 for MAE and 39.60 for MSE. In terms of low illumination \& Gaussian noise, AudioCANNet also beats CANNet, obtaining the MAE of 27.30 and MSE of 44.81 when $R=0.2, B=50$, which decreases by 17.6\% and 23.1\%. Interestingly, AudioCANNet also obtains lower MAE (12.68 vs. 15.41) and MSE (27.07 vs. 28.96) on high-quality images, therefore, in the case of using high-quality images, audio could be still helpful or at least not significantly reduce the performance (see table \ref{tab:clean_img}), but introducing audio makes model more stable in different scenarios, such as lowering illumination or adding noise.


\section{Discussion}
\begin{table}[t]
    \centering
    \caption{Performance on different audio representations and model settings. Given a raw audio, we first extract its MFCC and then use LSTM or CNN to further refine the feature. The counting model is our AudioCSRNet.}
    \label{tab:mfcc}
    \scalebox{0.75}{
    \begin{tabular}{c|c|c|c|c|c|c|c|c|c|c|c|c}
        \hline
         \multirow{3}{*}{\tabincell{c}{Model}} &\multicolumn{2}{c|}{Low resolution} &\multicolumn{4}{c|}{Gaussian noise} &\multicolumn{4}{c|}{Low illumination\&Gaussian noise}  &\multicolumn{2}{c}{\multirow{2}{*}{\textbf{Avg. Score}}}\\
         \cline{2-11}
         &\multicolumn{2}{c|}{$128\times 72$} &\multicolumn{2}{c|}{$\sigma=25/255$} &\multicolumn{2}{c|}{$\sigma=50/255$} &\multicolumn{2}{c|}{$R=0.2, B=25$} &\multicolumn{2}{c|}{$R=0.2, B=50$} \\
         \cline{2-13}
         &MAE $\downarrow$ &MSE $\downarrow$ &MAE $\downarrow$ &MSE $\downarrow$ &MAE $\downarrow$ &MSE $\downarrow$ &MAE $\downarrow$ &MSE $\downarrow$ &MAE $\downarrow$ &MSE $\downarrow$ &MAE $\downarrow$ &MSE $\downarrow$ \\
         \hline
         MFCC+LSTM & 20.64 & 37.30 & 24.83 & 43.36 & 25.78 &43.50 &38.33 &64.34 &43.88 &71.35 & 30.69 & 51.97\\ \hline
         MFCC+CNN &16.44 &29.83 &16.01 &32.70 &16.06 &31.56 &24.85 &46.73 &23.34 &41.98 &19.34 &36.56 \\
         \hline
    \end{tabular}
    }
\end{table}
Recall the questions that we want to answer (see Section~\ref{sec1}), we discuss in this section.

First, combining visual and auditory information is able to benefit crowd counting, in particular in the scenarios of low illumination, strong noise, low resolution and occlusion. If the image quality is high, introducing audio could result in comparable performance, since the visual information is good enough for counting. Another advantage of jointly applying auditory and visual cues is that the robustness of models in different scenarios can be enhanced. Models that only rely on vision could fail in some extreme scenarios, by contrast, combining auditory and visual information is able to handle the extreme scenarios.

Second, when illumination degrades, the performance of vision-based models could dramatically decrease (see Fig. \ref{diff-illumination}), whereas the audiovisual counting model is capable of obtaining lower MAE and MSE. Interestingly, the gap between vision-based and audiovisual models enlarges with the decrease of illumination and when there is no visual information, audiovisual counting models are able to estimate the number of people by only ``hearing'' the ambient sound. In terms of occlusion, audio is also helpful, in particular for the models that are sensitive to occlusion, such as CSRNet. And similar trend occurs in the scenario of occlusion---the gap enlarges with the increase of occlusion rate. Moreover, for low resolution images, combining audio and vision performs relatively well.

Third, in this paper we investigate in linear feature fusion approach, since it is simple and performs relatively well on other multi-modality tasks such as visual question answering (VQA) \cite{perez2018film}. And for the combination of vision and audio, linear feature fusion still performs well and in most cases audiovisual counting models outperforms their counterparts that only use visual information. In the future, we will pay more attention to how to fuse audio and vision for crowd counting and scene understanding.

Last but not least, we investigate in using different audio representations in AudioCSRNet and the results are shown in table~\ref{tab:mfcc}. Obviously, using CNNs to extract high-level features from spectrogram and MFCC is better than using LSTM. The possible reason is that CNNs consist of much more parameters than LSTMs, thus more powerful. Interestingly, MFCC could obtain similar performance compared with using spectrogram (see table~\ref{tab:lowq_img}), however, the computation of MFCC is slightly complex.

\section{Conclusion and Outlook}
In this paper, we investigated a novel audiovisual task, that imposes audio information for assisting visual crowd counting in extreme conditions.
We developed an audiovisual crowd counting dataset to facilitate progress in this field, which covers different scenes in different illuminations. Meanwhile, a feature-wise fusion model was developed to achieve audiovisual perception for crowd counting. Extensive experiments were conducted to explore audio effects in different visual conditions. We found that introducing audio is able to benefit crowd counting, in particular in the extreme conditions, such as low illumination, strong noise, low resolution and occlusion.

In the future, three directions should be considered. First, what is the best way to fuse audio and vision? In this paper, we studied a simple fusion approach, and some other methods should be investigated in the future. Second, does audio benefit crowd counting in other conditions, such as cross-scene? Third, if audio and video for the same scene are collected in different positions, does audio still benefit crowd counting?
\clearpage
%
%
\bibliographystyle{splncs04}
\bibliography{egbib}

\clearpage
\section{Appendix}
\renewcommand{\thetable}{S\arabic{table}}
\renewcommand{\thefigure}{S\arabic{figure}}

\begin{table}[h]
	\centering
	\caption{Performance on enhanced images.} \label{tab1}
	\begin{tabular}{c|c|c|c|c}
		\hline
		\multirow{2}{*}{Model} & \multicolumn{2}{c|}{$R=0.2, B=25$}  & \multicolumn{2}{c}{$R=0.2, B=50$} \\
		\cline{2-5}
		& MAE $\downarrow$ & MSE $\downarrow$ & MAE $\downarrow$ & MSE $\downarrow$ \\
		\hline
		CSRNet &33.83 &55.59 &43.06 &67.11 \\
		\hline
		AudioCSRNet &25.25 &46.21 &33.00 &55.06 \\
		\hline
	\end{tabular}
\end{table}

Since we lower the illumination and add Gaussian noise to the images, a naive approach for crowd counting is to first enhance the corrupted images and then feed them into the counting models. In our experiments, we use $R=0.2$ and $B=25, 50$ to randomly corrupt the input images and then we employ a Gaussian filter with $11\times 11$ kernel size and the standard deviation is adaptive to the size of kernel. After that we employ histogram equalization approach to alleviate the effect of lowering illumination.

Table \ref{tab1} shows the performance on enhanced images. Although the input images are enhanced, MAE and MSE scores are much higher than using high-quality images, e.g., CSRNet obtains MAE of 12.73 and MSE of 26.99 (see table 1 in our paper) using high-quality images, while it obtains 43.06 and 67.11 for MAE and MSE using enhanced images. Compared with using corrupted images (see table 2 in our paper), CSRNet performs better using enhanced images, e.g., it obtains MAE and  MSE of 33.83 and 55.59 using enhanced images, by contrast it obtains 35.78 for MAE and 62.76 for MSE using corrupted images with $R=0.2, B=25$. Also, when $R=0.2, B=50$, using enhanced images leads to better performance, e.g., 43.06 vs. 45.88 for MAE and 67.11 vs. 75.40 for MSE. Interestingly, in terms of AudioCSRNet, using enhanced images results in slightly worse performance than using corrupted images based on MAE, e.g., 25.25 vs. 25.06 and 33.00 vs. 27.33 for $B=25$ and $B=50$, respectively. The possible reason is that, after enhancing, people in images become more recognizable, in particular, the effect of low illumination is mitigated. Comparing AudioCSRNet to CSRNet, introducing audio significantly improves the performance, e.g., AudioCSRNet obtains 25.25 for MAE and 46.21 for MSE, while CSRNet obtains 33.83 and 55.59 for MAE and MSE when $R=0.2, B=25$.

In addition, we show more qualitative results in the following two figures.

\begin{figure*}[h]
\centering
\includegraphics[width=\textwidth]{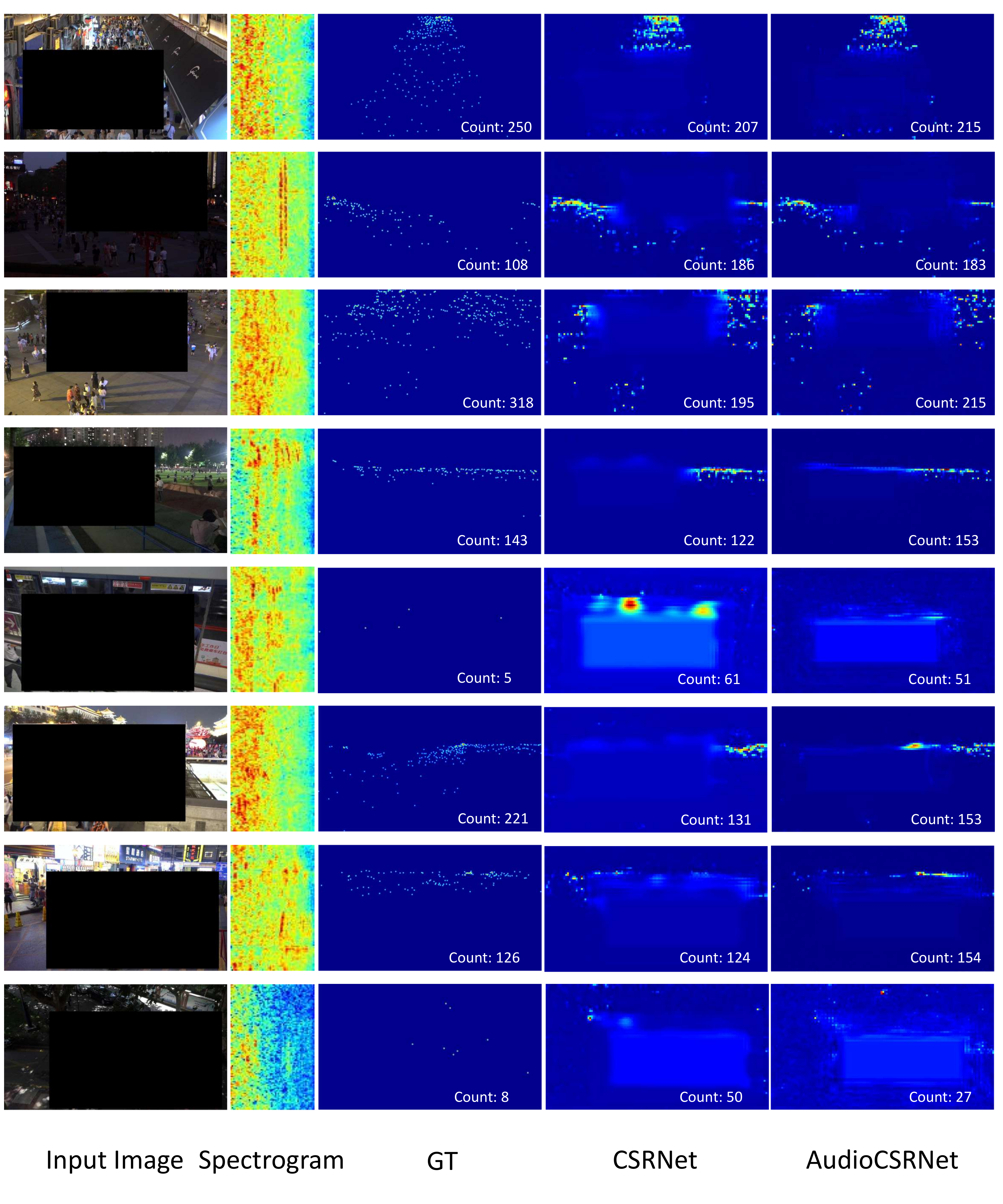}
\caption{Generated density maps on occluded images. In the case of occlusion, it is difficult to predict an accurate density map, in particular for the occluded area. In most cases, introducing audio is able to predict a more accurate number of people (see the numbers in the figure), which indicates that audio may not help us to localize people to generate accurate density maps when the images are in extremely low quality, but it is able to benefit counting the number of people.}
\end{figure*}

\begin{figure*}[h]
\centering
\includegraphics[width=\textwidth]{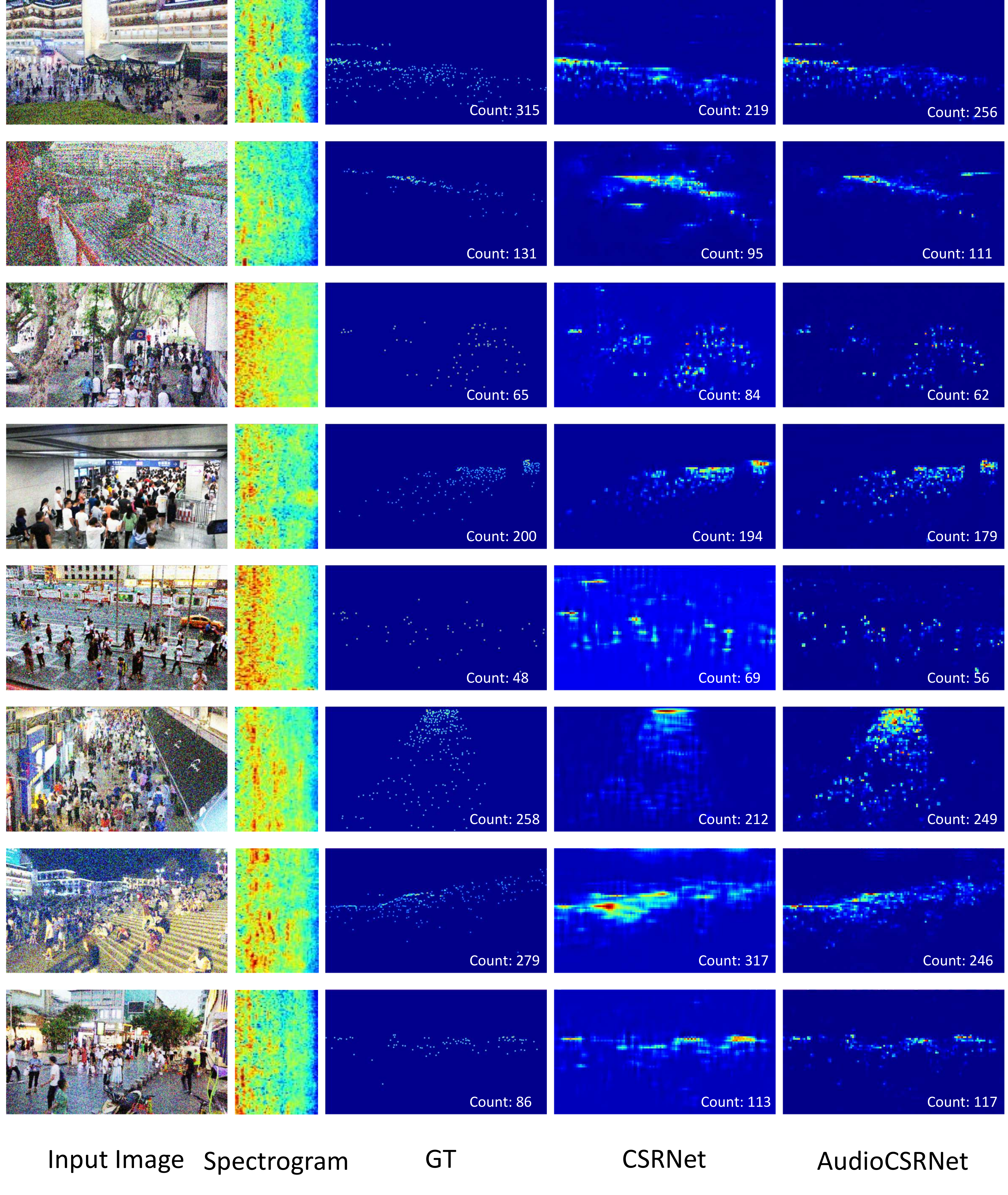}
\caption{Generated density maps on enhanced images. Although we enhance the images after Gaussian noise corruption and lowering illumination, the quality of the input images is still low in most cases. AudioCSRNet is able to generate high-quality density maps on low-quality images, however, when people in high-quality images are recognizable, AudioCSRNet could perform slightly worse (see the examples shown in the fourth and last rows).}
\end{figure*}

\end{document}